\documentclass[conference]{IEEEtran}
\IEEEoverridecommandlockouts
\usepackage{cite}
\usepackage{amsmath,amssymb,amsfonts}
\usepackage{algorithmic}
\usepackage{graphicx}
\usepackage{textcomp}
\usepackage{xcolor}
\usepackage{url}
\def\BibTeX{{\rm B\kern-.05em{\sc i\kern-.025em b}\kern-.08em
    T\kern-.1667em\lower.7ex\hbox{E}\kern-.125emX}}

\begin{document}
	
\urlstyle{same}

\title{MultiStream: A Simple and Fast Multiple Cameras Visual Monitor and Directly Streaming \\
{\footnotesize}
\thanks{}
}

\author{\IEEEauthorblockN{1\textsuperscript{st} Jinwei Lin}
\IEEEauthorblockA{\textit{Shenzhen Research Institute of Big Data} \\
Shenzhen, China \\
0000-0003-0558-6699}
}

\maketitle

%===========================================================
\begin{abstract}
Monitoring and streaming is one of the most important applications for the real time cameras. The research of this has provided a novel design idea that uses the FFmpeg and Tkinter, combining with the libraries: OpenCV, PIL to develop a simple but fast streaming toolkit MultiSteam that can achieve the function of visible monitoring streaming for multiple simultaneously. MultiStream is able to automatically arrange the layout of the displays of multiple camera windows and intelligently analyze the input streaming URL to select the correct corresponding streaming communication protocol. Multiple cameras can be streamed with different communication protocols or the same protocol. Besides, the paper has tested the different streaming speeds for different protocols in camera streaming. MultiStream is able to gain the information of media equipment on the computer. The configuration information for media-id selection and multiple cameras streaming can be saved as json files. \textnormal{\emph{(Abstract)}}

\emph{Monitoring; streaming; multiple; cameras; FFmpeg (key words)}

\end{abstract}

%===========================================================
\section{Introduction}
\label{sec:introduction}

The real-time camera is considered as one of the most important implementations and applications of video streaming, especially in the field of real time video monitoring. In the current situations of camera applied communications, video streaming has played an increasingly important role. Besides the streaming speed and efficiency, the convenience of the operation has been drawing more attention. In most of the implementations with a requirement of a simple and fast video streaming, a convenient toolkit or framework that is easy to operate is more available and reasonable. In most of the rapid testing situations, a simple but fast directly streaming framework is also significant for experiments.

There are two available methods to meet the requirements mentioned above that stream the real-time video. One is using the FFmpeg framework to establish the application or just using the command line to achieve the streaming, which is weak in the repeatability of the general situations or difficult in coding and weakening in visualization. Therefore, a visible and general toolkit developed on the ffmpeg is expected. The other method is developing a visible application that uses the abstraction and encapsulation of FFmpeg, such as VCL. However, even with the VLC (multimedia processor framework), there are some shortcomings on the implementations of camera video streaming (e.g.: diverse performances of streaming of application versions ). In terms of the interfaces of running the communications with other applications, VLC has provided some useful libraries on multiple platforms or language interactions. But it is not the direct interaction and communication with Python, which means VLC is not the preferred framework or toolkit for  a simple and fast implementation of camera streaming that is developed by Python. Moreover, if streaming function is the core component of the implementation, a small and fast streaming toolkit will be more available. In the field of simultaneous management, monitoring and streaming of multiple cameras, there is no efficient implementation of VLC.

The research of this paper has provided a novel design method to develop a streaming toolkit that is able to implement the simultaneous management, monitoring and streaming of multiple cameras. Combined with the corresponding functions of OpenCV, Tkinter and FFmpeg, the streaming toolkit is designed to satisfy the requirement of a simple but fast multi-streaming function. The code source of the research of this paper is open source on GitHub:\url{https://github.com/JYLinOK/FreedrawingDetecting}.

%Do not use any additional Latex macros.
%===========================================================
\section{Literature Review}
\label{sec:literature}

%------------------------------------------------------------------------- 
\subsection{Camera Real Time Streaming}

The technologies of the real time streaming of cameras is applied in various situations. With the advantage of the wireless video data communication, the video real time streaming is useful when it comes to the situations that need the real time video images from far away to make further managements \cite{2020Real}. The speed and the performance of real time video is the chasing goal of video streaming. To ensure the continuity and quality of the video stream, some corresponding communication protocols were developed, for instance, the $UDP, TCP, RTP, RTSP, HTTP$ and $RTMP$. The research of this paper will focus on the implementations of the video streaming of  $UDP, TCP, RTP$, which can be streamed directly without the configuration of establishment of streaming server.

%------------------------------------------------------------------------- 
\subsection{Streaming Camera with FFmpeg}

FFmpeg is considered as one of the most popular streaming frameworks and is open source. As a complete and cross-platform framework, FFmpeg is used as the fundamental component of various multimedia players and media information and data processors \cite{2020Multi}. Using the FFmpeg as the basic component, it will be easy to develop various cross-platform media processing applications. In terms of streaming the video, FFmpeg can stream the camera video whose images are real time with camera, and stream the video file, the streaming is also real time but without the communication with camera. The real time streaming with the communication with camera will have a higher requirement in implementation. To address the issues of multimedia processing, ffmpeg has provided the $dll$ libraries for third-party developed, such as $libavutil, libavcodec, libavformat, libavdevice, libavfilter$, $libswscale$ and $libswresample$. There are two methods to apply the FFmpeg, one is combining the corresponding libraries of FFmpeg into the developing applications, the other is using the command line to directly call the $ffmpeg$ sub-component of FFmpeg to implement the development.

%------------------------------------------------------------------------- 
\subsection{Streaming Camera with VLC}

VLC is a popular multimedia player and processing framework, which is open source and free. VLC is developed in FFmpeg and other excellent components, whose most common use is used as a multimedia and stream media player and streaming processor \cite{2018Real}. Similar to FFmpeg, VLC has also provided some libraries for third-party further developments, for instance, $libVLC, DVBlast$ and $biTStream$. There are also various streaming applications or multimedia processing applications that are designed based on the VLC. Compared with FFmpeg, VLC has provided the visible operation GUI which makes the operations and configurations more easy and direct. But VLC is weak in Python native developing combination and multiple cameras simultaneou streaming, monitoring and configuration, which are the issues that will be focused on and handled with in this paper. For some sample volume applications, the size of  VLC extensible components is too large, which means a simple and fast streaming and monitoring toolkit will be more appropriate.

%------------------------------------------------------------------------- 
\subsection{Multiple Cameras Monitoring with OpenCV}

The multiple cameras monitoring is useful and general when it comes to the situations that need the real time monitoring of multiple scenes. Using the real time multiple windows display to monitor or gain the visible image data from multiple scenes in different viewpoints is suggested \cite{2019Modeling}. The method to implement this set goal with FFmpeg or VLC is opening the application multiple times to gain the multiple application windows, which is complex and needs a long time to operate. Even when multiple windows are opened, the configuration of the layout of those windows is also complex. Therefore,  a better method to address this issue is to combine those multiple camera display sub-windows into one main window, which means using an application GUI to display multiple windows at the same time to satisfy the monitoring requirement. OpenCV is considered as a popular and useful open source development library to develop camera applications \cite{2012Real}. For the convenience of the communication with Python, using the Tkinter and its canvas control to build the GUI, and using the OpenCV to gain the real time video image data by capturing the images from the cameras is reasonable.

%===========================================================
\section{Research Methodology}
\label{sec:methodology}

%------------------------------------------------------------------------- 
\subsection{Main Functions and Layout Design}

\begin{figure}
	\centering
	\includegraphics[width=0.46\textwidth]{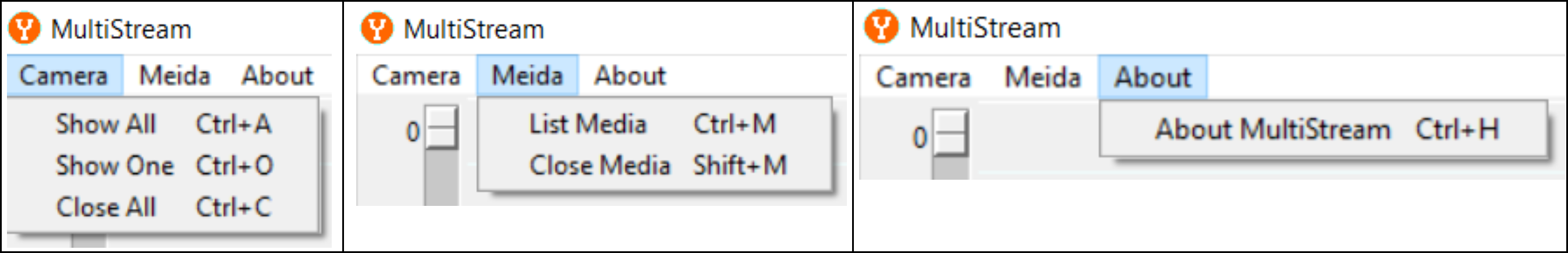}
	\caption{The main menu layout design of MultiStream.}
	\label{fig1}
\end{figure}

A reasonable layout design for the application is important to improve the user experience and the convenience of the toolkit. Developed on Tkinter, the GUI of MultiStream is easy to understand and handle. As shown in Figure \ref{fig1}, there are three key buttons on the main menu of MultiStream: $Camera, Media$ and $About$. The content box of the $Camera$ button is about the operations with the camera. The $Show All$ button is clicked  to display the videos of all of the existing cameras on the computer. The $Show One$ button is clicked to display one specific camera video customized. The $Close All$ button is clicked to close all of the camera video windows and restart the toolkit. The content box of the $Media$ button is about the operations with  media equipment on the computer. The $List Media$ button is clicked to list all of the media equipment on the computer with an information box, including the video and audio equipment. The $Close Media$ button is clicked to close the information box. The content box of the $About$ button is about displaying the relative information of MultiStream. The $About$ button is clicked to display the relative introduction information message box of MultiStream. Those buttons  on the menu are designed to make a simple but robust and concise GUI implementation of MultiStream.

%------------------------------------------------------------------------- 
\subsection{Customized Media Equipment Configuration}

\begin{figure}
	\centering
	\includegraphics[width=0.45\textwidth]{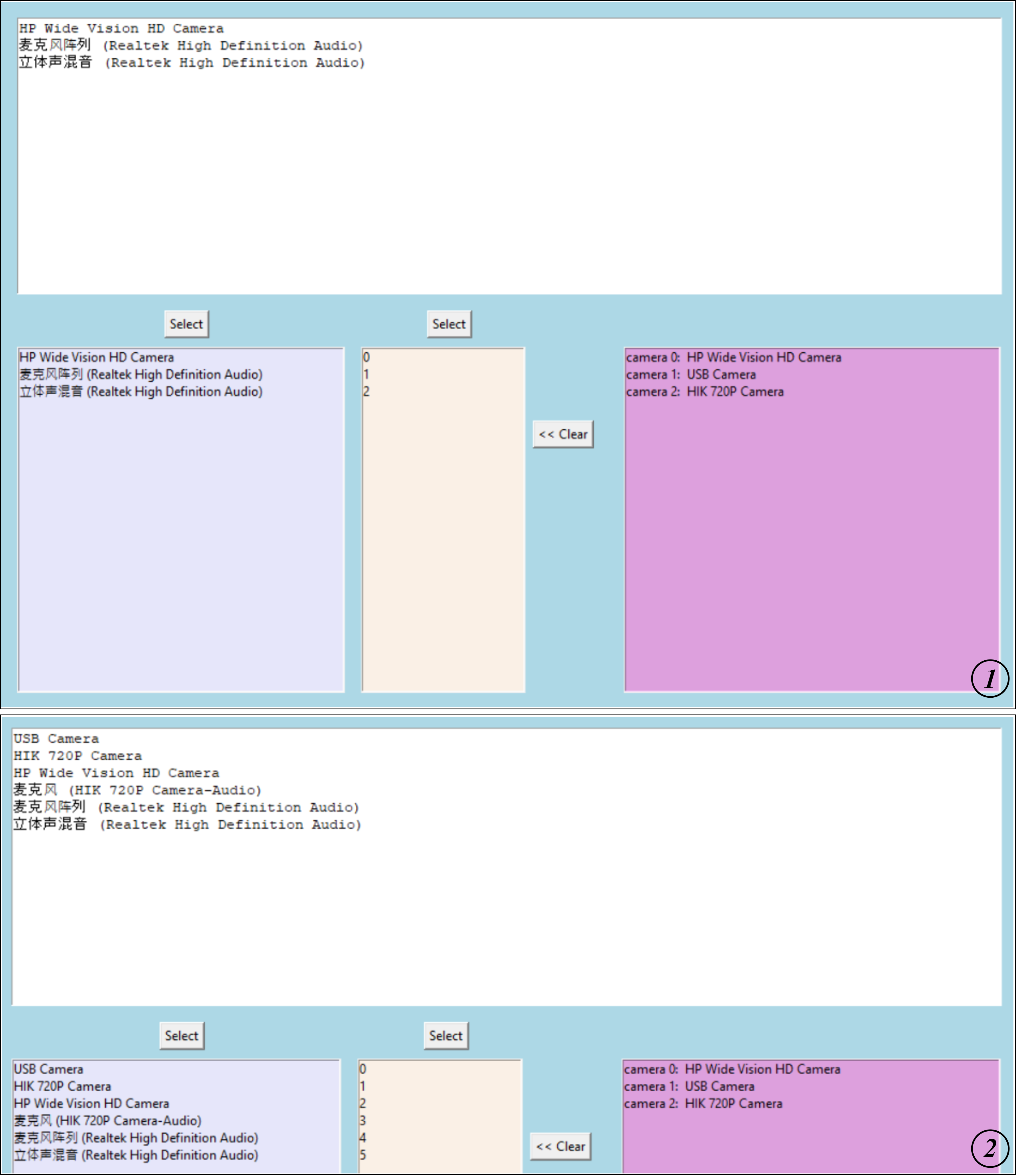}
	\caption{The main menu layout design of MultiStream.}
	\label{fig2}
\end{figure}

The fundamental process to operate the camera is to gain the specific information of each camera connected to the computer. To achieve this idea, using the OpenCV only is not enough, because of the limitation of python-opencv library, the python-opencv that is used in the implementation of this research is unable to get the corresponding name of each camera directly which is required in using the FFmpeg to stream the specific camera. Therefore, the name of each camera is the ID to identify each camera. In this research, to address this issue, using the name set gained from the FFmpeg, with the ID set gained from OpenCV, followed by the customized configuration of the camera by the user. As shown in Figure \ref{fig2}, each time clicking the button $List Media$, the current information of the inserted cameras on the computer will be updated and displayed. The sub graph $1$ is different with sub graph $2$ due to the change of the media equipment.

When it comes to the customized configuration operation, the user needs to click one of the names on the down left box followed by clicking the above  $Select$ button. Subsequently, clicking one the IDs on the down middle box followed by clicking the above $Select$ button. Following this, the content of the done right will be updated and shown automatically. The previous typing of the default name of the media equipment will remain, such as Chinese. The configuration information will be saved as a $json$ file to read in the next running time. If there is a configuration json file, the analyzer will load and read the information of the json file as a quick reference and remaining configuration. The following are the key codes to implement the corresponding algorithms:

\begin{verbatim}
def list_media():
   ...
   for i, item in enumerate(media_list):
           listbox_media.insert(i, item)
           listbox_cameras.insert(i, i)
       with open(json_path, 'r', 
       encoding='utf-8') as f:
           f_dict = json.load(f)
           if f_dict != {}: 
              media_cameras_dict = f_dict
   for i in media_cameras_dict:
           listbox_config.insert(i, 
           'camera ' + str(i) + ':  ' 
           + media_cameras_dict[i])
\end{verbatim}

%------------------------------------------------------------------------- 
\subsection{Automatically Detecting and Layout of Cameras Windows}

\begin{figure}
	\centering
	\includegraphics[width=0.46\textwidth]{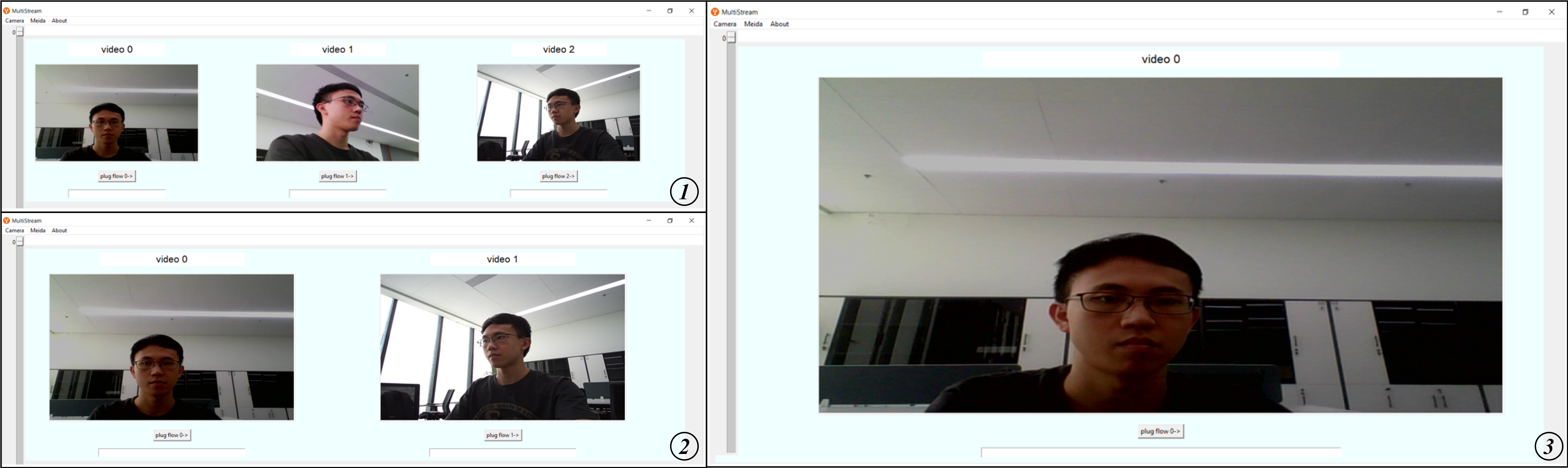}
	\caption{Automatically detect and resize of the layouts of camera windows.}
	\label{fig3}
\end{figure}

For a general multiple camera monitoring application, the camera is changeable. Therefore, due to the limited visual area of the display screen, a reasonable layout of the root window is important to achieve an available and useful GUI. To address this issue, the author of this paper has developed an automatic layout mechanism for MultiStream. As shown in Figure \ref{fig3}, when there are three cameras inserted into the computer, the display result of clicking the button $List Media$ is shown as subgraph $1$. When there are two cameras inserted into the computer, the display result of clicking the button $List Media$ is shown as subgraph $2$. When there is only one camera inserted into the computer, the display result of clicking the button $List Media$ is shown as subgraph $3$. Note that the max number of the cameras on one row is able to change in the code. If the count of the cameras is more than 3, the exceeding cameras will be shown in a new row. When the camera is displayed out of the screen, using the scrollbar on the left side to move the visible area. The detail parameters of the scrollbar can be also changed in the source code of this research.

The following is the key code to implement the design of this section mentioned above:

\begin{verbatim}
…
if cameras_num <= column:
   image_width = int((screenwidth 
   * adjust) / cameras_num)
   image_height = int(image_width
   * adjust_w_h)
else:
   image_width = int((screenwidth * adjust) 
   / column)
   image_height = int(image_width * 
   adjust_w_h)
…
for i in show_cameras:
   canvas = Canvas(frame2, bg = 'white', 
   width=image_width_local, height=
   image_height_local) 
   lable = Label(frame2, text = 'video '
   +str(i), bg='white', font=("consle", 16), 
   width=int(image_width/20), height=1)
   entry = Entry(frame2, width=int(
   image_width/10))
   button = Button(frame2, text = 
   'plug flow '+str(i)+'->', command=
   lambda index=i, len_cams=len_show_cameras:
   plugFlow(index, len_cams))
   
canvas_dict['canvas'+str(i)] = canvas
            row_i = int(i / column)
            column_i = i % column
…
\end{verbatim}

%------------------------------------------------------------------------- 
\subsection{Monitoring and Streaming Individual Specific Camera}

\begin{figure}
	\centering
	\includegraphics[width=0.4\textwidth]{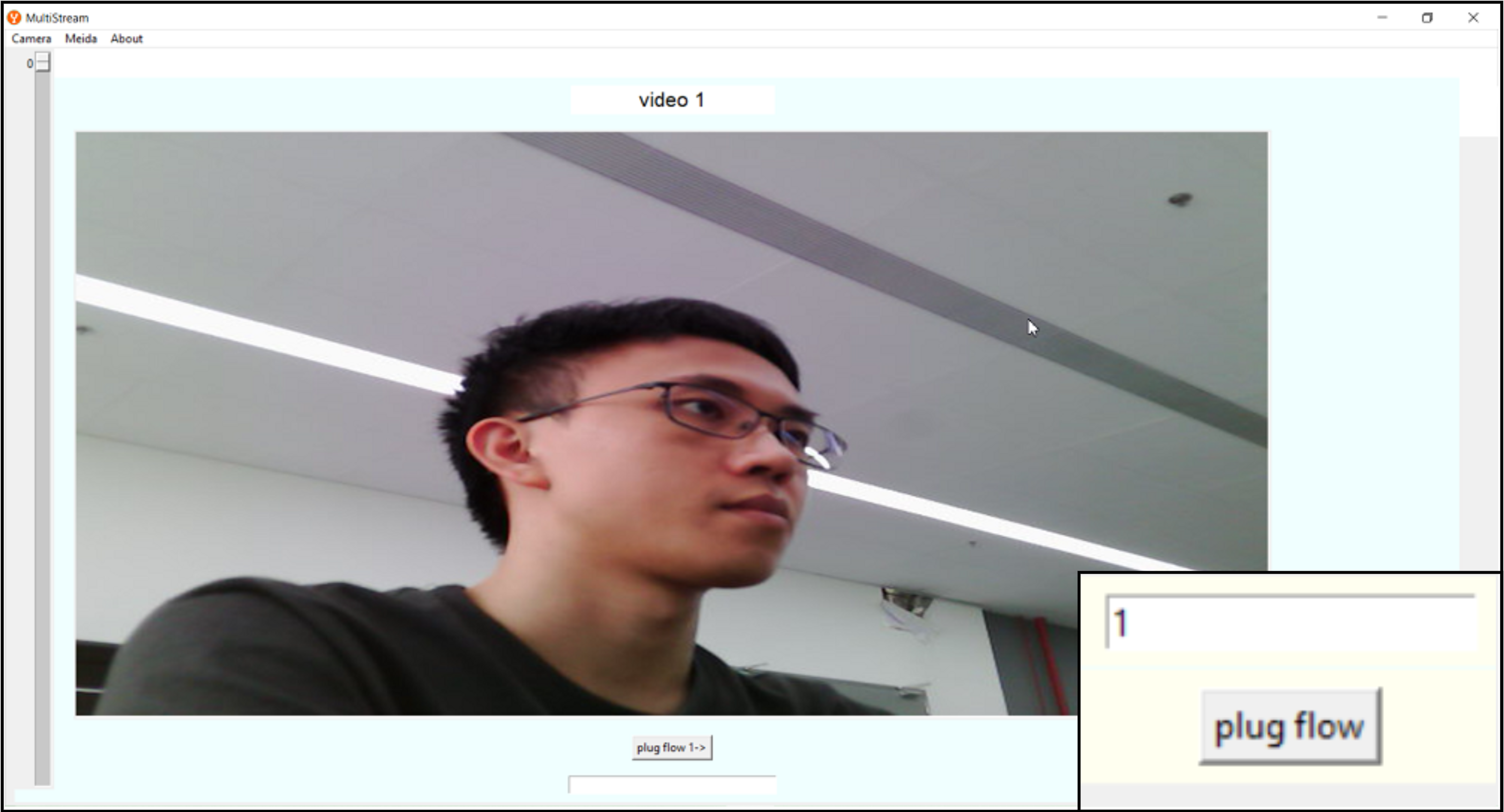}
	\caption{Monitoring and displaying one individual specific camera.}
	\label{fig4}
\end{figure}

Sometimes it is needed to monitor or stream the real time video of one specific camera in some specific individual situations. In this research, the function of the button $Show One$ is designed to monitor one individual camera. Besides the individual monitoring, the individual streaming of cameras is also important to implement. As shown in Figure \ref{fig4}, in MultiStream, there is an input textbox and streaming button $plug flow$ on the bottom of the sub-window of each camera. After the user inputting the url of the $UDP$, $RTP$ or $TCP$ and clicking the button $plug flow$, the core analyzer and interpreter of MultiStream will run the corresponding streaming program of the specific camera, following by the image of the camera streamed will be static.

\begin{figure}
	\centering
	\includegraphics[width=0.46\textwidth]{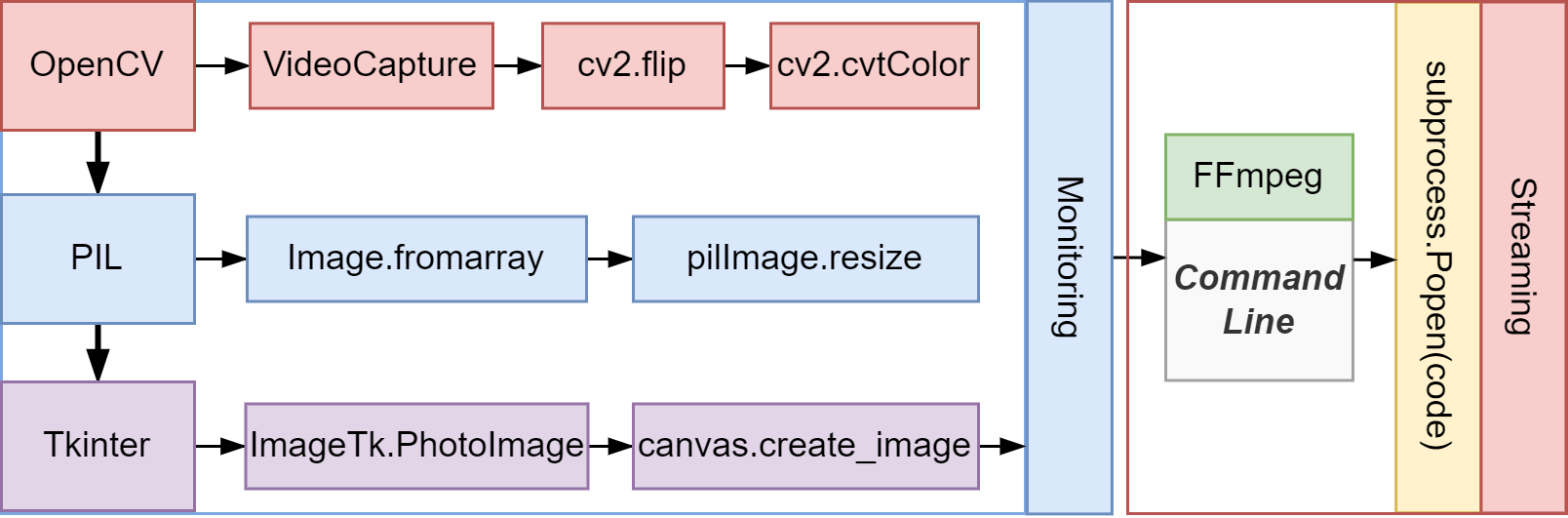}
	\caption{The basic principle of monitoring and streaming one camera.}
	\label{fig5}
\end{figure}

The implementation of the monitoring and streaming of one individual camera is the basic component of the algorithm of the implementation of the monitoring and streaming of multiple cameras processing algorithm in the research of this paper. As shown in Figure \ref{fig5}, there are three Python libraries used in this algorithm (i.e.: $OpenCV$, $PIL$ and $Tkinter$). Using the $VideoCapture$ function to gain the real time images data of the camera, followed by using the $cv2.flip$ to flip a 2D array from the last step. Subsequently using the function $cv2.cvtColor$ to change the image array from the $BGR$ color space to $RGBA$ color space. To the next section, the processed data will be sent to $PIL$, using the function $Image.fromarray$ to generate the corresponding image from the array, followed by using the function $pilImage.resize$ to change the showing shape of the image. Eventually, the processed image will be sent to Tkinter code, using the function $ImageTk.PhotoImage$ to change the image to a Tkinter image, followed by sending the Tkinter image to the Tkinter canvas control. Subsequently, using the function $canvas.create\_image$ to implement the display of monitoring.

During the display process of the monitoring, the user can click the button $plug flow$ in any time to run the processing of streaming the camera video. The main processing procedure of streaming the camera is, firstly, the code running monitor will receive the event calling of the button. Subsequently, the analyzer will analyze and gain the important input parameters of the button event, followed by generating the ffmpeg code with the format of the common line. Then the program will run the $subprocess. Popen(code)$ to run the ffmpeg command line code of streaming to start the streaming process.

%------------------------------------------------------------------------- 
\subsection{Monitoring and Streaming Multiple Cameras}

\begin{figure}
	\centering
	\includegraphics[width=0.46\textwidth]{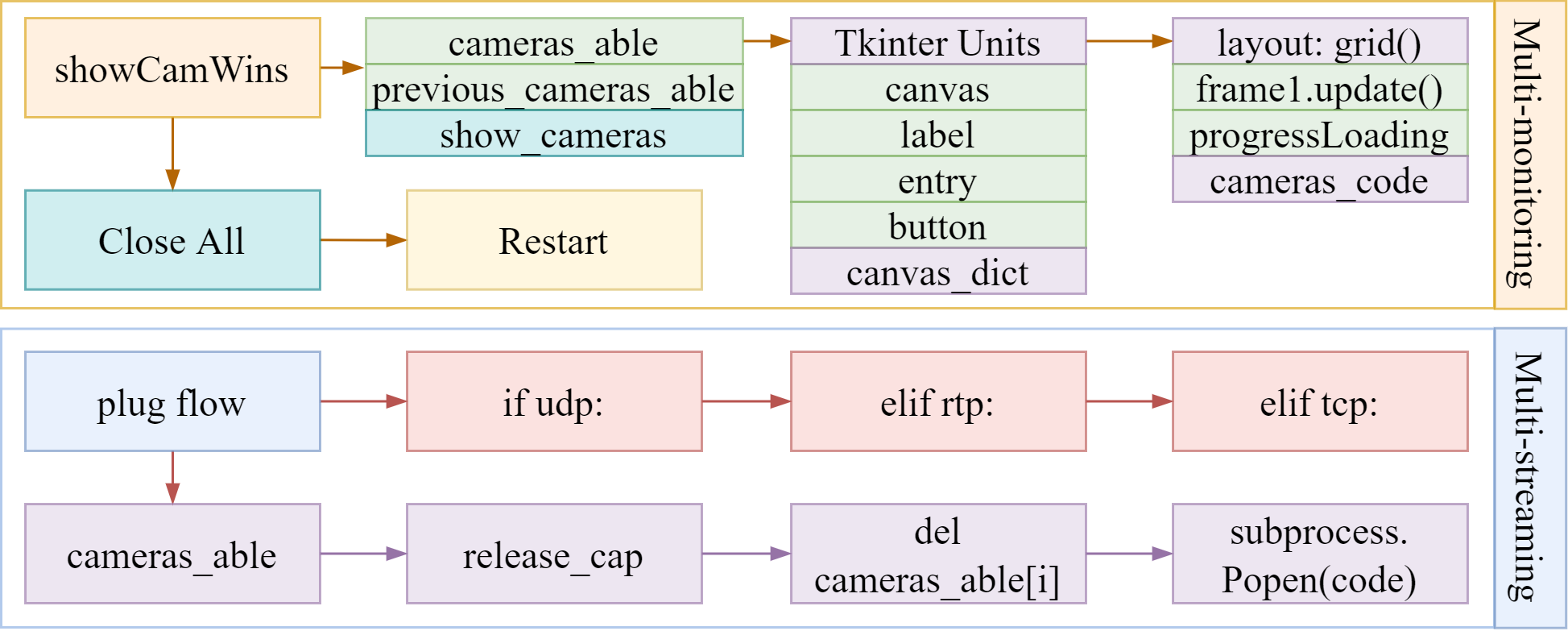}
	\caption{Monitoring and streaming of multiple cameras simultaneously.}
	\label{fig6}
\end{figure}

The design and implementation of the monitoring and streaming of multiple cameras simultaneously is the core of the research of this paper. Based on the design and development of the previous section, the design and implementation of the monitoring and streaming of multiple cameras simultaneously will be more available. As shown in Figure \ref{fig6}, in the design of the multiple monitoring of cameras, the function $showCamWins$ is the key implementation method. Firstly, the function $showCamWins$ will get a current list of the cameras to show video by the analysis of two global variables: $cameras\_able$ and $previous\_cameras\_able$, followed by gaining the local variable $show\_cameras$ for further analysis. Subsequently, the parameters will be sent to a program to build the layout with Tkinter. A Tkinter display unit consists of four basic components: $canvas$, $label$, $entry$, $button$. The processing result of the canvas will be saved in the list $canvas\_dict$. Eventually, using the Tkinter to set the configuration of the layout, and updating the display of the $frame 1$, followed by gaining the layout code from function $cameras\_code$. Then the final display of the monitoring multiple cameras will show.

The implementation principle of streaming multiple cameras is similar to the implementation principle of streaming one individual camera. Firstly, the user needs to enter the streaming URL in the entry box that is located in the bottom of each camera monitoring unit box. Subsequently, the analyzer of MultiStream will run the analysis to judge which communication protocol the user wants the camera to be streamed. The available  communication protocols to stream in the research are: $UDP$, $RTP$, $TCP$. With experiments, the speed correlation of streaming for these three protocols have a relationship of: $UDP>RTP>TCP$. Therefore, for the real time streaming on camera video, the most rapid  protocol to stream is using the $UDP$. To avoid the calling blocking caused by two processes calling the same camera simultaneously, it is necessary to release the camera before streaming it. In this research, the IDs of the cameras that are able to be streamed will be saved in the global variable $cameras$\_$able$, before the camera is going to be streamed, the ID of this camera will be deleted. Subsequently, the specific camera will be streamed by the function $subprocess.Popen(code)$. Each time the user clicks the $plug flow$ button will stream one camera video, with the input URL string in the corresponding entry box. Multiple cameras can be streamed simultaneously by continually clicking more cameras. Due to the layout design and running principle of Tkinter, when it comes to clicking multiple $plug flow$ buttons to stream multiple cameras, the right clicking sequence is from large ID to the small. Due to the memory preemption of running multiple processes to run multiple streaming, if there is unresponsive after clicking the $plug flow$ button, more times clicking will address this issue.  

The key code to implement the design mentioned above if shown as following:

\begin{verbatim}
…
while True:
   global release_cap
   if len(release_cap) > 0:
       for item in cap_dict:
          item_i = int(item[item.index('p')
          +1:])
          if item_i in release_cap:
             cap = cap_dict['cap'+
             str(item_i)]
             if cap.isOpened():
                cap.release()
             
   if len(cap_dict) > 0:
      cade_c = cameras_code(cameras_able)
      exec(str(cade_c))
      root_window.update()
      root_window.after(1)
…
\end{verbatim}

%===========================================================
\section{Experiments and Analyses}
\label{sec:experiments}

%------------------------------------------------------------------------- 
\subsection{Streaming multiple cameras with different protocols}

\begin{figure}
	\centering
	\includegraphics[width=0.47\textwidth]{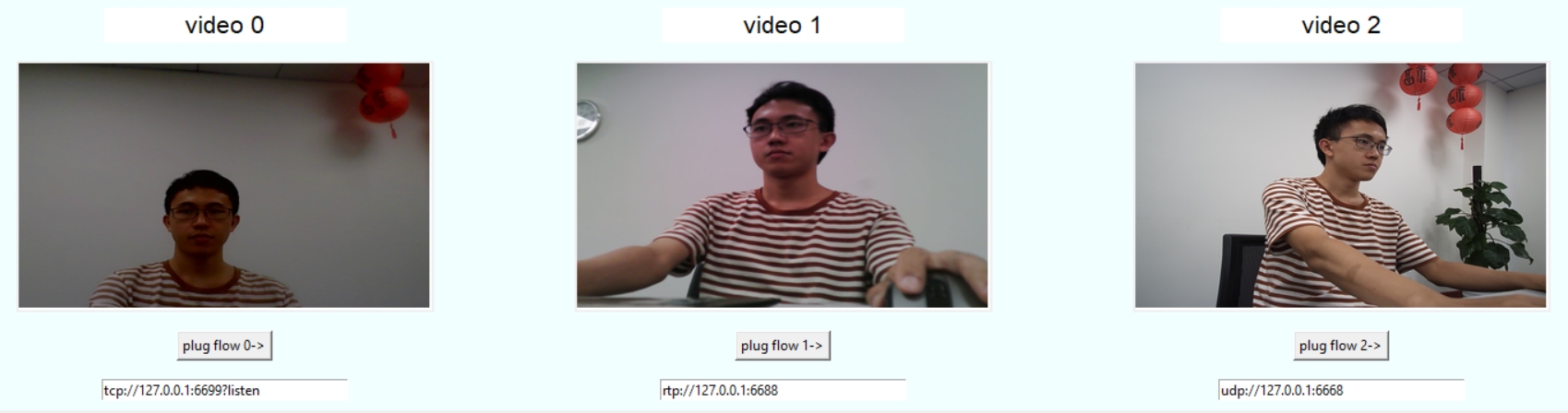}
	\caption{Streaming of multiple cameras simultaneously with different communication protocols: $UDP$, $RTP$ and $TCP$.}
	\label{fig7}
\end{figure}

\begin{figure}
	\centering
	\includegraphics[width=0.47\textwidth]{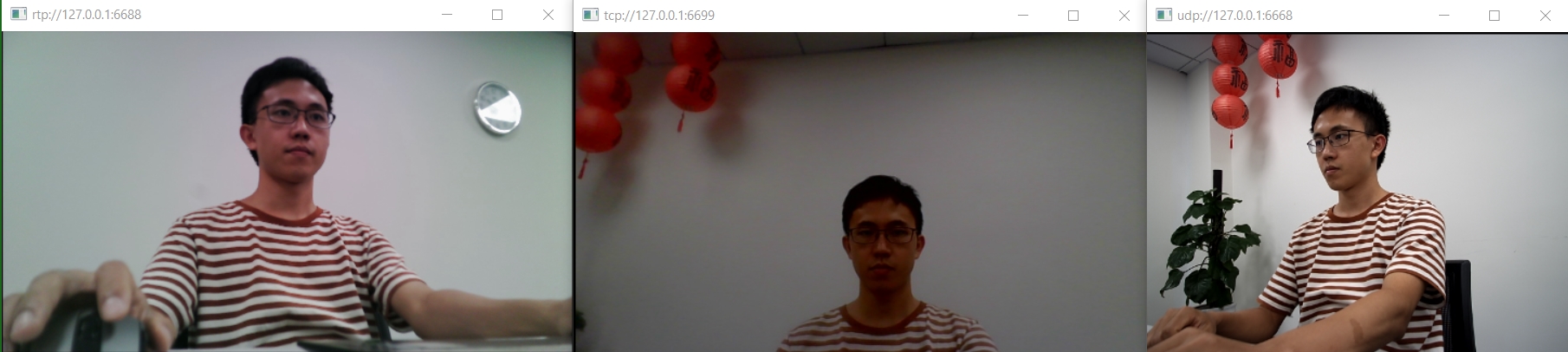}
	\caption{Streaming results of multiple cameras simultaneously with communication protocol: $UDP$, $RTP$ and $TCP$. Played with $ffplay$ to test.}
	\label{fig8}
\end{figure}

To ensure the performance of MultiStream, the research has implemented a number of experiments. Firstly, as shown in Figure \ref{fig7}, using MultiStream to stream the multiple cameras simultaneously with communication protocol: $UDP$, $RTP$ and $TCP$ individually is available and simple. When the streaming button of a certain camera is clicked, the real time image of this camera will pause. As shown in Figure \ref{fig8}, after the streamings of multiple cameras are successful, using the $ffplay$ to play the stream is feasible. By tests, the protocol $UDP$ has the most rapid streaming speed. Therefore, for the normal camera video streaming, to ensure the performance of streaming, the protocol $UDP$ should be the first choice.

The command line codes to play the streaming mentioned above of  $UDP$, $RTP$ and $TCP$ individually are shown as following:

\begin{verbatim}
…
ffplay -x 500 -y 375 udp://127.0.0.1:6668

ffplay -x 500 -y 375 -protocol_whitelist 
"file, udp, rtp" -i rtp://127.0.0.1:6688

ffplay -x 500 -y 375 tcp://127.0.0.1:6699
…
\end{verbatim}

%------------------------------------------------------------------------- 
\subsection{Streaming multiple cameras with a same protocol}

\begin{figure}
	\centering
	\includegraphics[width=0.47\textwidth]{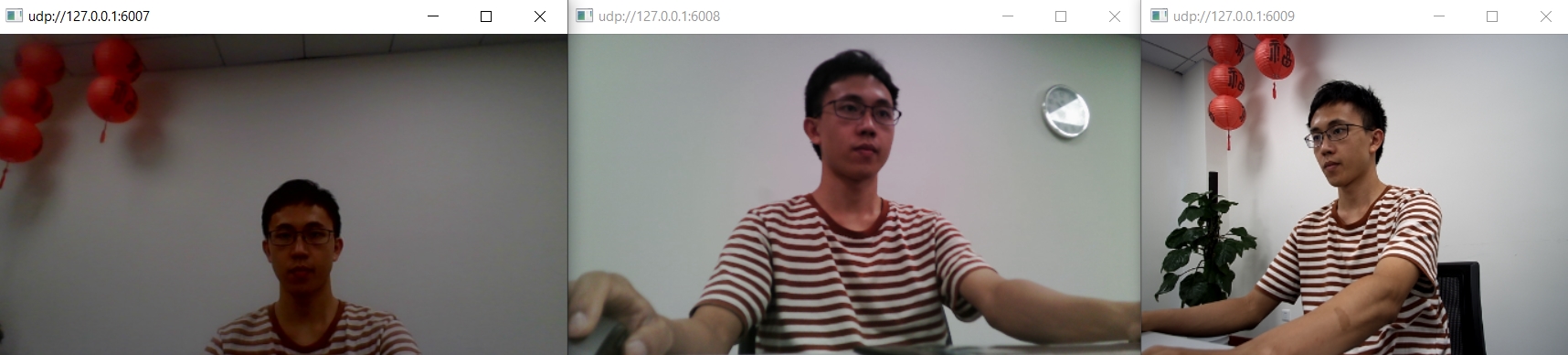}
	\caption{Streaming results of multiple cameras simultaneously using the same communication protocol and the streaming URL address is different. Played with $ffplay$ to test.}
	\label{fig9}
\end{figure}

Sometimes it is needed to stream multiple cameras with the same streaming protocol, such as $UDP$. It is supported to directly stream multiple cameras with the same streaming protocol in MultiStream. As shown in Figure \ref{fig9}, directly entering the different streaming URL in the different entry box of each corresponding camera and clicking the $plug flow$ button, the multiple cameras will be streamed with different URL and the same protocol.

After the series of experiments, it is proved that the function of streaming multiple cameras simultaneously with different communication protocols: $UDP$, $RTP$ and $TCP$ of MultiStream is available and useful. Compared with streaming with VLC or FFmpeg, MultiStream is more simple, visible and understandable to establish a new streaming. Besides, MultiStream supports streaming multiple cameras simultaneously and is more useful for practical applications. The json files of configuration of streaming for different communication protocols are saved in the path of: $"./config/plugflow.json"$.

%===========================================================
\section{Conclusion}
\label{sec:conclusion}

MultiStream is a useful tool kit for streaming one camera individually or streaming multiple cameras simultaneously. The operation of the streaming of MultiStream is visible, simple and understandable. Compared with VLC and FFmpeg, MultiStream has more advantages in streaming multiple cameras simultaneously with the visible control and streaming a large number of cameras repetitively with  one same configuration, which can improve the establishing time of streaming and decrease the time of repetitive configuration. For the small and rapid experiment or test, MultiStream is useful to establish an experiment platform. Developed by $ffmpeg$ instead of VLC, remains the excellent streaming performance of FFmpeg. Using the visible GUI can decrease the difficulties for users to operate. It is useful and meaningful to extend the support of more communication protocols of MultiStream in the future research, for instance, extending the $RTMP$ and $RTSP$ by importing the server design technologies.

%-------------------------------------------------------------------------
% bibtex
\bibliographystyle{IEEEtran} 
\bibliography{ref}

% biblatex with biber
% \printbibliography                

%-------------------------------------------------------------------------

\end{document}